\documentclass{llncs}
\usepackage{llncsdoc}

\usepackage{siunitx}
\usepackage{graphicx}
\usepackage{pdfpages}
\usepackage[T1]{fontenc}
\usepackage{subfig, siunitx, url}
\usepackage{amsmath, amssymb, bbm, nicefrac}
\usepackage{pifont}
\usepackage{color}
\usepackage{amsmath}

\usepackage[shadow,colorinlistoftodos,disable]{todonotes}%\usepackage{tikz}
\usepackage{xspace}	% for \xspace command

\newcommand{\competition}{RoboCup@Work\xspace}
\newcommand{\RC}{RoboCup\xspace}

\title{b-it-bots\\
	   \competition\\
	   Team Description Paper 2023}

\author{
Kevin Patel, Vamsi Kalagaturu, Vivek Mannava, Ravisankar Selvaraju, Shubham Shinde, Dharmin Bakaraniya, Deebul Nair, Mohammad Wasil, Santosh Thoduka, Iman Awaad, Sven Schneider, Nico Hochgeschwender and Paul G. Pl\"{o}ger
}
\institute{
		Hochschule Bonn-Rhein-Sieg\\
		Department of Computer Science\\
		Grantham-Allee 20, 53757 Sankt Augustin, Germany\\~\\
		Email: \texttt{<first\_name>.<last\_name>@inf.h-brs.de}\\
		Web: \url{https://www.h-brs.de/en/a2s/b-it-bots}
}

\begin{document}

\maketitle

\begin{abstract}
This paper presents the b-it-bots \competition~team and its current hardware and functional architecture for the KUKA youBot robot.
We describe the underlying software framework and the developed capabilities required for operating in industrial environments including features such as reliable and precise navigation, flexible manipulation, robust object recognition and task planning. New developments include an approach to grasp vertical objects, placement of objects by considering the empty space on a workstation, and the process of porting our code to ROS2.
\end{abstract}

\section{Introduction}
\label{section::introduction}

The b-it-bots \competition~team at the Hochschule Bonn-Rhein-Sieg (H-BRS) was established in the beginning of 2012. Participation in various international competitions has resulted in several podium positions, including first place at the world championship \RC 2019 in Sydney and second place at the online RoboCup Championship in 2021.
The team consists of Master of Science in Autonomous Systems students, who are advised by two professors.
The results of several research and development (R\&D) as well as Master's thesis projects have been integrated into a highly-functional robot control software system.
Our main research interests include mobile manipulation in industrial settings, omni-directional navigation in unconstrained environments, environment modeling and robot perception in general.

\section{Robot Platform}\label{section_platform}

\begin{figure}[t]
  \centering
  \includegraphics[height=0.7\textwidth]{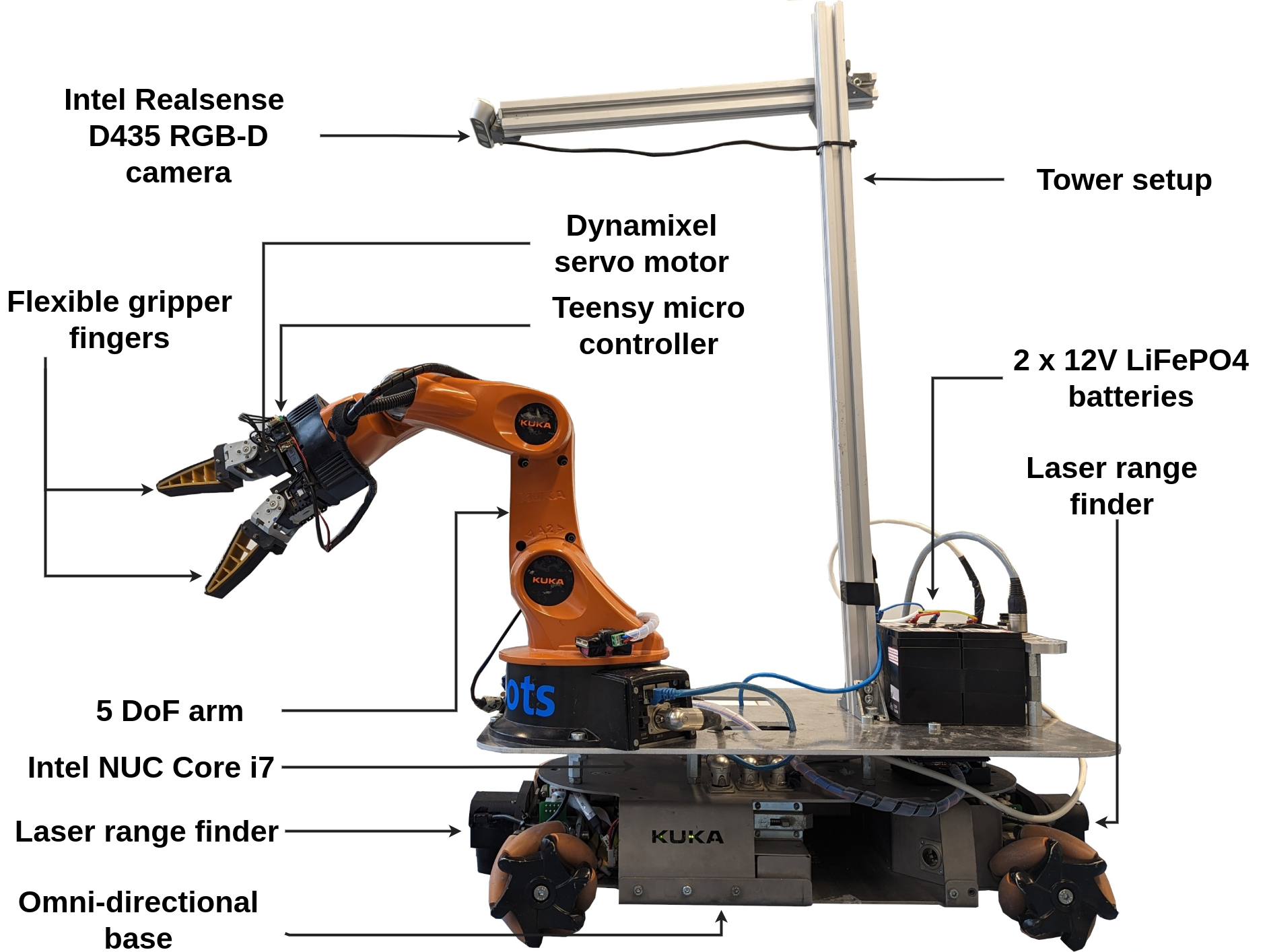}
  \caption{b-it-bots robot configuration based on the KUKA youBot~\cite{padalkar2019_robocup}}
  \label{fig:brsu_youbot}
\end{figure}

The KUKA youBot \cite{youbot_store} is the applied robot platform of our \competition~team (see Figure \ref{fig:brsu_youbot}).
It is equipped with a 5-DoF KUKA manipulator, a two finger gripper and an omni-directional platform. The standard internal computer of the youBot has been replaced by an Intel NUC\cite{intel_nuc} with an Intel Core i7 processor.
In the front and the back of the platform, two Hokuyo URG-04LX laser range finders are mounted to support robust localization, navigation and precise placement of the omnidirectional base.
Each laser scanner is configured with an opening angle of 190$^{\circ}$ to reduce the blind spot area to the left and right of the robot.

Over the years, we have experimented with different sensors and sensor configurations for perception-related tasks, including placing an RGB-D or RGB camera on the end-effector, or mounting a fixed RGB-D camera at a height above the arm.
Our current configuration is seen in Figure~\ref{fig:brsu_youbot}, which consists of a tower-mounted Intel Realsense D435 RGB-D camera, with the tower profile mounted to the metal base plate of the robot. Additionally, we may also mount a second camera (of the same model) between the gripper fingers. 
The different configurations represent a trade-off between getting a full view of the workspace during perception and better close-up views of the objects during manipulation.
But in general, this sensor information is used for vital perception tasks such as 3D scene segmentation, object detection and recognition and barrier tape detection.

\begin{figure}[t]
  \centering
  \includegraphics[height=0.45\textwidth]{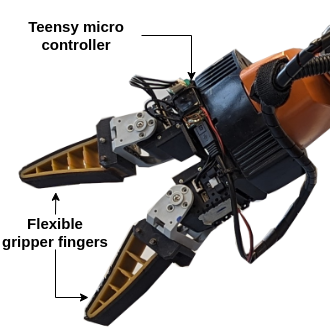}
  \caption{Flexible-finger gripper and teensy micro controller~\cite{teensy}}
  \label{fig:teensy-zoom-view}
\end{figure}

The youBot gripper has undergone further customization, where it now features flexible fingers actuated by Dynamixel AX-12 servo motors. These motors offer both position control and force-feedback information, and are controlled by a Teensy microcontroller which is shown in Figure~\ref{fig:teensy-zoom-view}. The robot's new configuration is powered by 2 lightweight LiFePO4 batteries, which offer a significant improvement compared to the previous battery system.

All technical drawings to the previously described modifications, as well as various 3D printed sensor mounts for the laser scanner and the RGB-D camera etc., have been made public \cite{brsu_technical_drawings}.

\section{Robot Software Framework}\label{section_architecture}
The software framework of the system is based on ROS (Robot Operating System) \cite{Quigley2009}. Information is passed between functional components through ROS communication channels, specifically using topics. The use of topics allows for non-blocking communication and the ability to monitor communication between nodes at any time. The broad range of tools offered by ROS are employed for visualizing, testing, and debugging the entire system. Our development approach emphasizes the creation of small, lightweight, and modular components that can be repurposed in different processing pipelines, including those in different domains or on alternative robot platforms, such as the Care-O-bot 3 \cite{Reiser2009}.
We have also standardized our nodes with the addition of \texttt{event in} and \texttt{event out} topics.
Our components listen to the \texttt{event in} topic which expects simple command messages and allow for: starting, stopping or triggering (run once) of nodes. The components provide feedback of their status on the \texttt{event out} topic when they finish.
This allows us to coordinate and control the components with either simpler state machines or task planning; in either case, the control flow and data flow between the components remains separated. This also allows us turn off computationally expensive nodes when they are not needed.

We have begun porting our software to ROS2~\cite{macenski2022robot}, since the last version of ROS1 will reach its end-of-life by 2025.
Most of our code is open source at \cite{public_atwork_repository}.

\section{Navigation}
Several components have been developed and integrated to move the robot from one place to another in cluttered and even narrow environments.

\subsection{Map-based Navigation}
The navigation components we use are based on the ROS navigation stack \emph{move\_base} which uses an occupancy map together with a global and local path planner.
For the local path planner a Dynamic-Window-Approach (DWA) is deployed which plans and executes omni-directional movements for the robot's base.
This enhances the maneuverability, especially in narrow environments. \\

The vast amount of configuration parameters of the \emph{move\_base} component have been fine-tuned through experiments with several and differently structured environments in simulation (e.g. a corridor, narrow passages, maze, etc.).

\section{Perception} \label{section_perception}
Several components have been developed for processing the image and point cloud data from the tower-mounted camera.
\subsection{Object Recognition}
\begin{figure}
  \centering
  \includegraphics[width=1.0\textwidth]{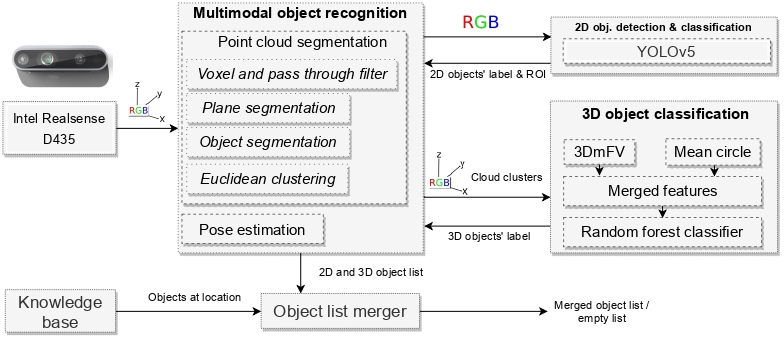}
  \caption{Object perception pipeline~\cite{padalkar2019_robocup}}
  \label{fig:perception-pipeline}
\end{figure}

Perception of objects relevant for industrial environments is particularly challenging.
The objects are typically small and often made of reflective materials such as metal.
We use an RGB-D camera which provides both intensity and depth images of the environment.
This enables effective scene segmentation and object clustering.
But the spatial resolution is low even at the close range, and a significant degree of flickering corrupts the depth images.
Thus, for the object detection and recognition task we use both 3D and 2D methods. The perception pipeline is outlined in Figure \ref{fig:perception-pipeline}.\\

For 3D perception, we capture a single point cloud and downsample it using a voxel grid filter in order to reduce the computational complexity. We then apply passthrough filters to restrict the FOV which removes irrelevant data and further reduces the computational burden. In order to perform plane segmentation, we first calculate the surface normals of the cloud and use a sample consensus method to segment a single horizontal plane. The convex hull of the segmented plane is computed and represented as a planar polygon. The prism of points above the polygon are segmented and clustered to individual object pointclouds.

For object recognition, we use a combination of features described in~\cite{Thoduka2016} and the 3D modified Fisher vector~\cite{ben20183dmfv} to train a random forest classifier. \\

In our approach to 2D perception, we employ YOLOv5~\cite{jocher2020zenodo} for object detection, with the inferencing implemented in C++ using the Open Neural Network Exchange (ONNX)~\cite{bai2019} model.
The object pose and grasping axis are then determined by using a Principal Component Analysis (PCA) approach on the point cloud within the 2D bounding box.

The results from 3D and 2D perception are combined along with information about the objects from the referee box to make a final classification result.

\subsection{Cavity Recognition}
For the Precision Placement Task, the robot is required to insert objects into cavities.
The correct cavity has to be chosen and the respective object needs to be precisely placed into it.\\
The cavities in the Precision Placement Test are detected by applying Canny edge detection and contour detection and classified using a YOLOv5~\cite{jocher2020zenodo} network.

\subsection{Rotating Table Test}
For the Rotating Table Test, the 3D positions of objects are tracked using a nearest neighbour approach after applying 3D segmentation to the pointcloud.
Based on the measured velocities and positions of the objects, the predicted arrival time of the target object is calculated and eventually grasped using background change detection with the camera pointed at the table~\cite{padalkar2019_robocup}.

Due to the inaccurate estimation of the movement of the objects on the table, the robot currently often misses objects. To address this problem, we have introduced 2D object tracking~\cite{scharf2022} to the perception pipeline. The fast and reliable tracking performed on each frame received from the camera would result in obtaining full trajectories of the objects of interest. This can be used to obtain a more accurate estimation of the object's future positions.

Currently, we have no data available for training models for 2D object tracking, thus our solution is based on the concept of tracking-by-detection, where a dataset for training object detectors is sufficient. Two methods considered by us are: Tracktor~\cite{bergmann2019tracking} and DeepSORT~\cite{wojke2017simple}. 

Tracktor~\cite{bergmann2019tracking}\footnote{ \url{https://github.com/phil-bergmann/tracking_wo_bnw}} firstly initializes the trajectory from the detection done by an object detector (such as Faster-RCNN or YOLOv5), secondly, the tracking is performed. The latter is a process that consists of two steps, namely: bounding box regression and bounding box initialization. The purpose of bounding box regression is to extend the trajectory to the new video frame. So the aim is to find the object's new position by reusing the position of the bounding box in the old frame. It is assumed that a high frame rate is provided, and the target object moves only slightly between frames. This assumption allows performing bounding box regression by applying ROI pooling on the features obtained from the new frame and bounding box coordinates from the previous frame. After the regression is done, it is decided which trajectories should be deactivated. The choice is based on the occlusions of the tracked objects. The bounding box initialization is used to find new targets in the frame for which the trajectory should be created. The difference from the first initialization (at the very beginning) is that here the new trajectory, for the given object, is started only when the IoU (Intersection over Union) with any of the existing trajectories is smaller than a certain threshold.

\begin{figure}
	\centering
	\includegraphics[width=1.0\textwidth]{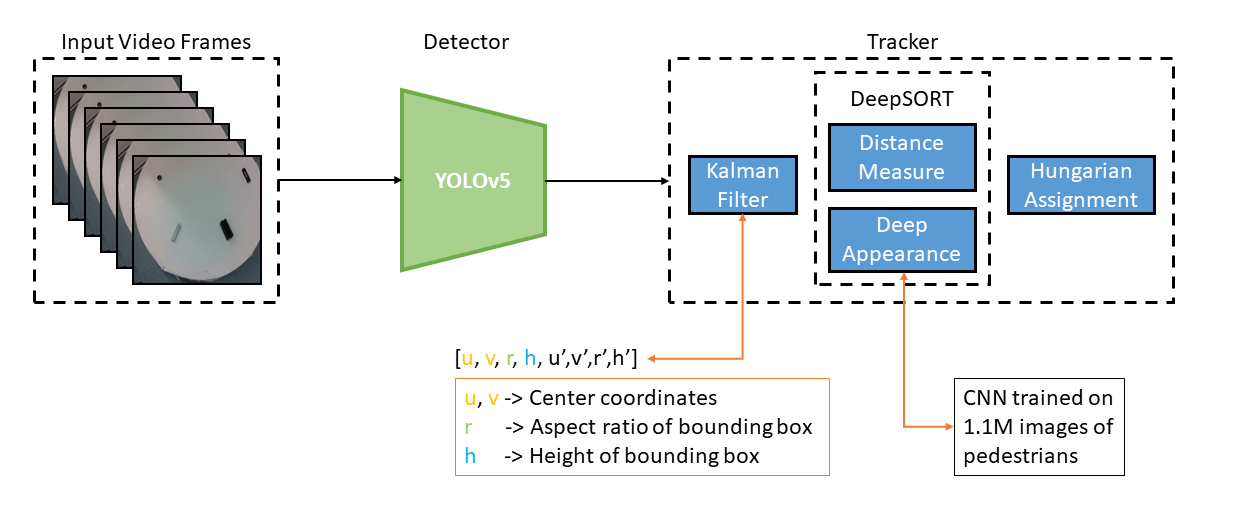}
	\caption{YOLOv5 + DeepSORT tracker architecture~\cite{wojke2017simple} }
	\label{fig:deepsort}
\end{figure}

We have also tested another solution for the purpose of tracking objects for the rotating table task, which uses the DeepSORT model \cite{wojke2017simple}\footnote{ \url{https://github.com/nwojke/deep_sort}} (Figure~\ref{fig:deepsort}). It is an improvement over the previous model, SORT \cite{bewley2016simple}. Whereas the previous model performed associations based on the overlap of the estimated bounding box generated using Kalman filter and the bounding box generated by the detector for the next frame, DeepSORT also incorporates appearance features extracted from the detection using a custom CNN architecture which outputs a 128-length feature vector as the output. The features of two detections are then compared using the cosine distance of the two vectors. It then combines the information from the motion features and the appearance features to associate objects using the Hungarian algorithm. It has shown improvement over the previous approach, primarily in reducing the number of ID switches. The model used has YOLOv5 detector as the backbone. More details of our approach can be found in~\cite{scharf2022}, and the code\footnote{\url{https://github.com/VincentSch4rf/rtt_tracking}} and dataset\footnote{\url{https://github.com/VincentSch4rf/RoboCup-RTT-Dataset}} are also available publicly.

\section{Object Manipulation}\label{section_manipulation}
In order to grasp objects reliably, several components have been developed and integrated on the robot. \\

From the object recognition component, the pose of the object to be grasped is retrieved. 
This pose is the input to a \textit{pre-grasp planner} component which computes a pre-grasp configuration based on the type of grasp, a distance offset and constraints imposed by the robot's manipulator and end effector. Due to its kinematic constraints, the robot might not be able to reach this computed pre-grasp configuration with its end effector.
Thus, a set of poses is sampled around the object's pose. An inverse kinematics solver is then used to find one reachable pre-grasp pose from the list of sampled poses. 
Objects laying horizontally on the workspace are approached from above.
For vertical objects (objects with a height that is more than a predefined value), we developed an approach to grasp them from the front. The robot positions itself sufficiently away from the object, such that the arm can approach the object from the front rather than from above.

Once the end effector reaches the grasp pose, the gripper of the robot is closed.
A grasp monitor checks whether the object is grasped successfully utilizing the force and position feedback of the two Dynamixel motors.

\subsection{Empty Space Placement}

In previous years, objects were placed on the target workstation using predefined arm configurations, without consideration for objects that may already be present on the workstation.
We have developed a new approach which perceives the workstation before placement and have integrated it into the working pipeline of the robot.
The previous method of object placement is reliable only when the height of the workstation is known, and no other objects are present.
Our current approach acquires this information dynamically and uses point cloud data from the RGB-D camera to find a plane for placement.
Poses are sampled in the free space of the plane, such that a minimum distance is maintained to other perceived objects on the plane, and the object is placed at one of the poses based on their reachability.
This approach allows the robot to place more objects on a workstation and it also dynamically adjusts to the variations on the workstation.
We are currently experimenting with placement on uneven surfaces.

\section{Task Planning} \label{sec:TaskPlanning}
Many robot application, especially in competitions, have been developed using finite-state machines (FSM).
But even for apparently simple tasks, such a FSM can be very complex and thus become easily confusing for humans.
Therefore, our current FSMs have been replaced with a task planner. \\

We test using both the Mercury 2014 planner \cite{katz2014mercury} and the LAMA planner \cite{richter2011lama}.
The LAMA planner is built on the Fast Downward planning system and uses PDDL. As such, it uses similar interfaces to those of the Mercury planner. The planners allow specifying various cost information. In terms of \competition, these costs and can be, for example, distances between locations or probabilities of how good a particular object can be perceived or grasped. \\

Small and clear state machines covering basic actions, like \texttt{move-to-location}, \texttt{perceive-object}, \texttt{grasp-object} or \texttt{place-object} are used as actions for the planner.
For a particular task, the planner then generates a sequence of those actions in order to achieve the overall goal. Finally, this plan is being executed and monitored.
In case of a failure during one of the actions, replanning is being triggered and a new plan is generated based on the current information available in the \emph{knowledge base}.

\section{Conclusion}\label{section_conclusion}
In this paper we presented several modifications applied to the standard youBot hardware configuration as well as the functional core components of our current software architecture. 
Besides the development of new functionality, we also focus on developing components in such a manner that they are robot independent and can be reused for a wide range of other robots with even a different hardware configuration.
We applied the component-oriented development approach defined in BRICS \cite{Bischoff2010} for creating our software which resulted in high feasibility when several heterogeneous components are composed into a complete system.

\section*{Acknowledgement} 
\scriptsize
We gratefully acknowledge the continued support of the team by the b-it Bonn-Aachen International Center for Information Technology, Bonn-Rhein-Sieg University of Applied Sciences and AStA H-BRS.

%%% REFERENCES %%%
\bibliography{literature_references}
\bibliographystyle{unsrt}

%%% APPENDIX %%%

\end{document}